\documentclass{l4dc2026}

\usepackage{amsfonts, mathrsfs}
\usepackage{tikz}
\usetikzlibrary{matrix}
\usepackage{braids}
\usetikzlibrary{shapes}
\tikzstyle cross=[preaction={draw=white, -, line width=6pt}, thick]
\tikzstyle normal=[thick]
\usetikzlibrary{arrows,calc,through,backgrounds,matrix,decorations.pathmorphing}
\usetikzlibrary{shadows}
\usetikzlibrary{positioning, fit}

\usepackage{acro}

\DeclareAcronym{mdp}{short = MDP ,	long  = Markov decision process}
\DeclareAcronym{rl}{short = RL ,	long  = reinforcement learning}
\DeclareAcronym{pi}{short = PI ,	long  = policy iteration}
\DeclareAcronym{lspi}{	short =LSPI ,	long  = least-squares policy iteration}
\DeclareAcronym{lsvi}{	short =LSVI ,	long  = least-squares value iteration}
\DeclareAcronym{klspi}{	short =KLSPI ,	long  = kernel-based  least-squares policy iteration}
\DeclareAcronym{kae_lspi}{	short =KAE-LSPI ,	long  = Koopman autoencoder-based least-squares policy iteration}
\DeclareAcronym{kae_lstd}{	short =KAE-LSTD ,	long  = Koopman autoencoder-based least-squares temporal-difference}
\DeclareAcronym{lstd}{	short =LSTD ,	long  = least-squares temporal-difference}
\DeclareAcronym{klstd}{	short =KLSTD ,	long  = kernel-based least-squares temporal-difference}
\DeclareAcronym{td}{	short =TD ,	long  = temporal-difference}
\DeclareAcronym{kae}{	short =KAE ,	long  = Koopman autoencoder}
\DeclareAcronym{edmd}{	short =EDMD ,	long  = extended dynamic mode decomposition}
\DeclareAcronym{kedmd}{	short =kEDMD ,	long  = kernel-based extended dynamic mode decomposition}
\DeclareAcronym{ci}{	short =CI ,	long  = Confidence Interval}
\DeclareAcronym{uat}{	short =UAT ,	long  =Universal Approximation Theorem}
\DeclareAcronym{nn}{	short =NN ,	long  =neural network}
\DeclareAcronym{rbf}{	short =RBF ,	long  =radial basis functions}
\DeclareAcronym{ald}{	short =ALD ,	long  =approximate linear dependency}

\title[Feature identification in LSPI using the Koopman operator framework]{Automatic feature identification in least-squares policy iteration using the Koopman operator framework}
\usepackage{times}

\author{%
 \Name{Christian {Mugisho Zagabe}} \Email{christian.mugishozagabe@tu-dortmund.de}\\
 \addr Department of Computer Science \& Lamarr Institute, TU Dortmund University, Germany.
 \AND
 \Name{Sebastian Peitz} \Email{sebastian.peitz@uni-paderborn.de}\\
 \addr Department of Computer Science, Paderborn University, Germany.%
 }

\begin{document}

\maketitle

\begin{abstract}%
In this paper, we present a \ac{kae_lspi} algorithm for \ac{rl}. The \ac{kae_lspi} algorithm is based on reformulating the so-called \textit{least-squares fixed-point approximation} method in terms of \ac{edmd}, thereby enabling automatic feature learning via the \ac{kae} framework. The approach is motivated by the lack of a systematic procedure for selecting features or kernels in linear \ac{rl} techniques. Unlike previous approaches, no features or kernels need to be fixed \textit{a priori} in our method. 
We discuss the theoretical convergence of \ac{kae_lspi} under suitable assumptions, similar to \ac{klspi}. 
We empirically compare our method with two existing algorithms: \ac{lspi}  and \ac{klspi}. In particular, for the chainwalk problem,  the resulting policies are either optimal or nearly optimal for both three approaches. For the cart-pole system, \ac{kae_lspi} outperforms the other two approaches in terms of policy performance.
\end{abstract}
\acresetall
\acresetall

\section{Introduction}
\label{sec:introduction}

In \ac{rl}, an agent takes a (deterministic or probabilistic) action, which causes a probabilistic transition of the \ac{mdp} (i.e., the environment). This in turn delivers a reward to the agent.
The ultimate goal of the agent is to maximize the expected total reward by developing an optimal feedback strategy (policy). It is encoded in the so-called state-action value function, or $Q$-function which is defined via the Bellman equation. 
For simple \ac{rl} problems with finite state and action spaces and where the probability transition and reward function of the \ac{mdp} are known, tabular methods are useful \citep{sutton1998reinforcement}. However, for real-world applications, the agent's challenge is to learn the optimal policy by estimating the $Q-$function from a set of observed data samples.

Over the last decades, several least-squares-based value function approximation algorithms were developed.
They are motivated by convergence results and error bounds for gradient-based \ac{td} algorithms with linear approximation \cite[Chapter 9]{sutton1998reinforcement}. Their main architecture considers the value function as a linear combination of certain basis functions or features (polynomials, \ac{rbf}, kernels, etc.) that constitute the dictionary. They then aim to learn optimal parameters such that the approximated value function is sufficiently close to the true function. Extensions for control (\ac{lspi} and \ac{klspi}) were considered in \citet{lagoudakis2003least,xu2007kernel,huang2012novel}.
As illustrated and discussed in \citet{lagoudakis2003least}, the manual selection of the basis functions in \ac{lspi} algorithms considerably affects the results and no universal choice is available. This fundamental obstacle motivated the \ac{klspi} algorithm where the \ac{ald}-based kernel sparsification approach was employed \citep{xu2007kernel,huang2012novel}. In this technique, data samples and an \textit{a priori} fixed kernel function are used to construct a data dictionary in which feature vectors are obtained by evaluating the kernel at the data points and removing those that can be approximately represented as a linear combination of others. 
Despite its advantages, two main obstacles of the \ac{klspi} algorithm can be observed. As in the \ac{lspi} case, the first one is the manual choice of the fixed kernel function to define the data dictionary. 
The second obstacle is the uncontrolled number of features that are  dynamically learned.
The main objective of this article is to demonstrate that it is possible to learn features universally from data in \ac{lspi} contexts, without the need to define them beforehand or to use a fixed kernel. Our approach is not intended to replace current algorithms, which are sometimes simpler, more practical, or less computationally expensive.

The Koopman operator is a linear operator that describes the evolution of observable functions \citep{mauroy2020koopman}. In the data-driven context, the Koopman operator is commonly approximated by the so-called \ac{edmd} algorithm which enables the study of the dynamics of an unknown system from data. Similarly to \ac{lspi} in \ac{rl}, the \ac{edmd} algorithm requires an \textit{a priori} choice of a dictionary of basis functions which can considerably influence the performance of the algorithm. To overcome this problem, a \ac{kedmd} algorithm was developed in \citet{kevrekidis2016kernel}. 
Dictionary learning was proposed in \citet{li2017extended}.
Even better, \citet{otto2019linearly,azencot2020forecasting} proposed the \ac{kae} algorithm which is a machine learning approach to the dictionary. In this instance, a \ac{nn} allows one to generically learn a set number of features from data without the need for explicitly fixed basis functions or a kernel. Our fundamental concept is to use the \ac{kae} procedure in the framework of the linear value function approximation theory in \ac{rl}. Lastly, we note that the connection between the Bellman equation and the Koopman operator was used in settings other than our \ac{lspi}. For instance, we can cite \citet{rozwood2024koopman} in relation to entropy \ac{rl} algorithms, \citet{zanini2021data,zanini2022value} in relation to value-function estimation without control and policy gradient algorithms, etc.

In this study, to resolve the issues related to the manual selection of the basis functions in \ac{lspi} and the kernel in \ac{klspi}, we propose the \ac{kae_lspi} algorithm in which the dictionary of basis functions is learned from the data using the \ac{kae} algorithm. This algorithm is mainly based on reformulating the classical \textit{least-squares fixed-point approximation} method \citep{lagoudakis2003least} in terms of the Koopman operator framework. Therefore, we learn the dictionary from the data.
Finally, the obtained \ac{kae} dictionary is used in the classical \ac{lspi} context to solve \ac{rl} problems. 
The main contributions of this paper are as follows:
\begin{itemize}
	\item We reformulate the classical \ac{lspi} algorithm in \ac{rl} in terms of the Koopman operator framework.
	\item We introduce the \ac{kae}
	technique step in the \ac{lspi} algorithm  to learn the dictionary of basis functions from data, thereby avoiding their manual selection. 
	\item We discuss the convergence analysis of \ac{kae_lspi}.
\end{itemize}

In Section \ref{sec:preliminaries}, we briefly provide preliminaries on \ac{rl} and the classical \ac{lspi} algorithm, as well as the Koopman operator framework and the \ac{kae} technique. Our main algorithm is described in Section 	\ref{sec:main} and illustrated in Section \ref{sec:illustrations} using
two classes of examples. Section 	\ref{sec:conclusion} gives concluding remarks and
perspectives. Finally, convergence analysis of \ac{kae_lspi}  is discussed in Appendix~\ref{append:converg}.
\section{Preliminaries}
\label{sec:preliminaries}

\subsection{Markov decision processes} An \ac{mdp} is a \(5-\)tuple \((\mathcal{S},\mathcal{A},\mathcal{P},R,\gamma)\), where \(\mathcal{S}\) and \(\mathcal{A}\) are state and action spaces, \(\mathcal{P}\) is a Markovian transition model indicating the transition to state \(s'\) when taking action \(a\) in the state \(s\): \(\mathcal{P}(s'|s,a)\), \(R: (s,a,s')\in \mathcal{S}\times \mathcal{A} \times \mathcal{S} \to R(s,a,s')\in \mathbb{R}\) is the reward function and \(\gamma\in ]0,1[\) is the discount factor for future rewards.
The goal of the agent is to maximize its total expected reward by designing a (in our case deterministic) policy
\(\pi:s\in \mathcal{S} \to \pi(s) \in \mathcal{A}.\)
The expected, discounted, total reward when taking action \(a\) in state \(s\) by following policy \(\pi\) is encoded in the state-action value function (or \textit{\(Q\)-function})
\[Q^\pi(s,a)=\mathbb{E}_{a_t\sim \pi; s_t\sim \mathcal{P}}\left[\sum_{t=0}^{\infty}\gamma^tr_t|s_0=s,a_0=a\right],\]
and the optimal policy maximizes
the value function \(Q^\pi\).
\subsection{Bellman optimal equation}
For a state-action pair \((s,a)\), the exact optimal \(Q^\pi\) values follow the \textit{Bellman optimal equation}
{\small \begin{equation}\label{eq:bellman_optimal}
		Q^{*}(s,a)= \sum_{s'\in \mathcal{S}} \mathcal{P}(s'|s,a) \left[R(s,a,s')+\gamma \max_{\pi} Q^\pi(s',\pi(s'))\right].
\end{equation}}
In real-world settings one has access only to sampled data from the \ac{mdp} problem, or either \(\mathcal{S}\) or \(\mathcal{A}\) may be infinite sets which makes \eqref{eq:bellman_optimal} difficult to solve. To overcome these difficulties, \(Q\)-function approximation approaches have been developed. 

\subsection{Least-squares policy iteration methods}

In the \(Q\)-function linear approximation approach, the goal is to adjust the parameters \(w\) in order to have 
\begin{equation*}
	Q^\pi(s,a)		\approx \hat{Q}^\pi(s,a;w)=\sum_{j=1}^k \phi_j(s,a)w^\pi_j=\phi(s,a)w^\pi,
\end{equation*}
where \(\phi=\big(\phi_1 \,\phi_2\,\ldots\,\phi_k\big)\in \mathbb{R}^{1\times k}\) is an \textit{a priori} fixed dictionary of basis functions (or features). 	Therefore, the optimization problem \eqref{eq:bellman_optimal} is transferred to the parameters \(w^\pi\in \mathbb{R}^{k\times 1}\), i.e., 
\begin{equation}\label{eq:Q_optimal_approx}
	\hat{Q}^*(s,a)=\phi(s,a)w^*,
\end{equation}
where \(w^*\) represents the optimal parameter.

Using a given dataset \(\{s_i,a_i,r_i,s'_i\}_{i=1,\ldots, L}\), the \textit{least-squares fixed-point approximation} \citep{lagoudakis2003least} consists of solving the linear equation
\begin{equation}\label{eq:fixed_point_method}
	\mathbf{\Phi^\top} \left( \boldsymbol{\Phi}-\gamma \mathbf{\Phi'_\pi} \right) w^\pi = \mathbf{\Phi^\top R},
\end{equation}
where
{\small \begin{equation}\label{eq:phi_r}
		\boldsymbol{\Phi}=\left[\phi(s_i,a_i)\right]_{ i=1}^L
		\in \mathbb{R}^{L\times k},\,
		\boldsymbol{R}=\big(
			r_1\,\, 
			r_2\,\,
			\ldots\,\,
			r_L
		\big)^\top \in \mathbb{R}^{L\times 1}
\text{ and } 
	\mathbf{\Phi'_\pi}
	=\left[\phi\big(s_i',\pi(s_i')\big)\right]_{i=1}^L\in \mathbb{R}^{L\times k}
\end{equation}
with \(L\) the size of the dataset and \(k\) the number of the basis functions fixed \textit{a priori} in the dictionary \citep{lagoudakis2003least}.
One of the biggest obstacles in this approach is the manual selection of the dictionary \(\phi\). 

\subsection{Koopman operator approach}

At this point, we denote the state by \(x\) to be consistent with the notation \(x=(s,a)\) used in Section \ref{sec:main}.
Suppose that we have a stochastic dynamical system
\begin{equation}\label{eq:nonlinear_map}
	x'=F(x,\xi),
\end{equation}
described by the nonlinear transition map \(F\) on \(\mathcal{X}\subseteq\mathbb{R}^n\), where \(\xi\) represents a random variable \citep{colbrook2024beyond,nuske2023finite}. The map \(F\) induces an infinite dimensional linear operator \(\mathcal{K}_F\) acting on an observable function \(f\in \mathcal{F}=\{g:x\in \mathcal{X} \to g(x)\in \mathbb{R}\}\) on \(\mathcal{X}\) by :
\[(\mathcal{K}_F f)(x)=\mathbb{E}_\xi\left[f(F(x,\xi))\right]= \mathbb{E}_{x'\sim F(x,\xi)}\left[f(x')\right].\]
This leads to an infinite-dimensional linear system on \(\mathcal{F}\)
\begin{equation}\label{eq:koopman_infinite}
	f'=\mathcal{K}_F f.
\end{equation}

Although at this stage the function space \(\mathcal{F}\) is general, for the theoretical point of view  we will consider the function space \(\mathcal{F}=L^2(\mathcal{S}\times \mathcal{A}, \mu)\), where \(\mu\) is an appropriate measure (see Appendix~\ref{append:converg}).
The goal of the Koopman framework is to design a dictionary of features \(\phi=\big(\phi_1 \,\phi_2\,\ldots\,\phi_k\big)\in \mathbb{R}^{1\times k}\) where 
the nonlinear dynamics \eqref{eq:nonlinear_map} is lifted and the infinite-dimensional dynamics is approximated in a finite-dimensional subspace:
\begin{equation}\label{eq:koopman_finite}
	z'= z \boldsymbol{K}.
\end{equation}
Therefore, the matrix \(\boldsymbol{K}\in \mathbb{R}^{k\times k}\) represents an approximation of the Koopman operator \(\mathcal{K}_F\) on the subspace spanned by the features \(\phi_1,\phi_2,\ldots,\phi_k\). One popular approach is the \ac{edmd} algorithm \citep{williams2015data}:
\begin{equation*}
	\boldsymbol{K} = \arg\min_{\tilde{\boldsymbol{K}}\in \mathbb{R}^{k\times k}} \dfrac{1}{L}\sum_{i=1}^{L} \|\phi\big(x'_i\big) - \phi\big(x_i\big) \tilde{\boldsymbol{K}} \|^2,
\end{equation*}
where \(\{\big(x_i,x'_i\big)\}_{i=1,\ldots,L}\) are data pairs from the dynamics \eqref{eq:nonlinear_map}.

Many research works, among which we particularly mention the \ac{kae} technique \citep{azencot2020forecasting,otto2019linearly,champion2019data,lusch2018deep}, have been conducted to improve the \ac{edmd} algorithm. Instead of fixing the dictionary, the \ac{kae} uses a deep \ac{nn} to approximate the observable function $f$. 

The architecture of the \ac{kae} used in this work is represented in Figure \ref{fig:kae}. 
\begin{figure*}[t]
	\centering
	\begin{tikzpicture}[scale=0.95,
		node distance=0.8cm and 1.2cm,
		every node/.style={font=\small},
		neuron/.style={circle, draw, minimum size=0.6cm, inner sep=0pt, fill=white},
		input/.style={neuron, draw=red, thick},
		latent/.style={neuron, draw=blue, thick},
		hidden/.style={neuron, draw=black},
		annot/.style={text centered}
		]
		\node[input] (in-1) at (0, 0.6) {\(x_{1}\)};
		\node[input] (in-2) at (0, -0.6) {\(x_{2}\)};
		\node[draw, rectangle, gray, dashed, inner sep=6pt, fit=(in-1) (in-2), label=above:{\(x=(s,a)\)}] (xbox) {};
		
		\foreach \i in {1,...,5} \node[hidden] (h1-\i) at (2, -\i*0.7+2.1) {};
		\foreach \i in {1,...,4} \node[hidden] (h2-\i) at (4, -\i*0.7+1.75) {};
		
		\foreach \i in {1,2,3} \node[latent] (z-\i) at (6, -\i*0.8+1.6) {\(z_{\i}\)};
		\node[annot, above=0.2cm of z-1] {\(z=\Phi(x)\)};
		
		\draw[->, ultra thick] (6.7, 0) -- node[above] {\(\boldsymbol{K}\)} (7.7, 0);
		
		\foreach \i in {1,2,3} \node[latent] (znext-\i) at (8.4, -\i*0.8+1.6) {\(z'_{\i}\)};
		\node[annot, above=0.2cm of znext-1] {\(z'=z\boldsymbol{K}\)};
		
		\foreach \i in {1,...,4} \node[hidden] (h3-\i) at (10.4, -\i*0.7+1.75) {};
		\foreach \i in {1,...,5} \node[hidden] (h4-\i) at (12.4, -\i*0.7+2.1) {};
		
		\node[input, draw=red] (out-1) at (14.4, 0.6) {\(x'_{1}\)};
		\node[input, draw=red] (out-2) at (14.4, -0.6) {\(x'_{2}\)};
		\node[draw, rectangle, gray, dashed, inner sep=6pt, fit=(out-1) (out-2), label=above:{\(x'=(s',a')\)}] (xbox_out) {};
		
		\newcommand{\connectlayers}[2]{
			\foreach \a in {#1} \foreach \b in {#2} \draw[->, >=stealth, gray!40] (\a) -- (\b);
		}
		\connectlayers{in-1, in-2}{h1-1,h1-2,h1-3,h1-4,h1-5}
		\connectlayers{h1-1,h1-2,h1-3,h1-4,h1-5}{h2-1,h2-2,h2-3,h2-4}
		\connectlayers{h2-1,h2-2,h2-3,h2-4}{z-1,z-2,z-3}
		\connectlayers{znext-1,znext-2,znext-3}{h3-1,h3-2,h3-3,h3-4}
		\connectlayers{h3-1,h3-2,h3-3,h3-4}{h4-1,h4-2,h4-3,h4-4,h4-5}
		\connectlayers{h4-1,h4-2,h4-3,h4-4,h4-5}{out-1, out-2}
		
		\node[below=1.5cm of h1-3, font=\bfseries] {Encoder \(\Phi\)};
		\node[below=1.5cm of h4-3, font=\bfseries] {Decoder \(\Psi\)};
		\node[font=\bfseries] at ($(z-3)!0.5!(znext-3) + (0,-1.3)$) {Koopman};
		
		\node[text=red, scale=0.7, fill=white] at ($(in-1)!0.5!(h1-3)$) {Sigmoid};
		\node[text=red, scale=0.7, fill=white] at ($(h1-3)!0.5!(h2-2)$) {Sigmoid};
		\node[text=red, scale=0.7, fill=white] at ($(h2-2)!0.5!(z-1)$) {$\tanh$};
		\node[text=red, scale=0.7, fill=white] at ($(znext-1)!0.5!(h3-2)$) {Sigmoid};
		\node[text=red, scale=0.7, fill=white] at ($(h3-2)!0.5!(h4-3)$) {Sigmoid};
	\end{tikzpicture}
	\caption{ Architecture of the \ac{kae}.  
			In our experiments, the \(\tanh\) activation function is only used  between the last two layers of the encoder part; in the other hidden layers, we use the scaled sigmoid function: \(f(x) = 1/\big(1 + \exp(-2x)\big)\). 
            }
	\label{fig:kae}
\end{figure*}
The input data \(x\) (\(x=(s,a)\) in our case) are lifted via an encoder \(\Phi\) to a higher dimension where the dynamics represented by the matrix \(\boldsymbol{K}\) is linear. The output data \(x'\) are obtained from the feature space via a decoder \(\Psi\). 
The goal of the \ac{kae} is to minimize the weighted loss function 
		\begin{equation*}
L_{\text{tot}}=\lambda_{\text{red}}L_{\text{rec}}+\lambda_{\text{pred}}L_{\text{pred}}+\lambda_{\text{dyn}}L_{\text{dyn}}+\lambda_{\text{bell}}L_{\text{bell}}\\+\lambda_{\text{spar}}L_{\text{spar}}+\lambda_{\text{orth}}L_{\text{orth}},
				\end{equation*}
			where
			the reconstruction loss \(L_{\text{rec}}\) enforces recovery of the state \(x\): \[L_{\text{rec}}=\mathcal{L}_{\text{Huber}}\left(\Psi(z),x\right),\]
			with \[\mathcal{L}_{\text{Huber}}(a,b)=\begin{cases}\frac{(a-b)^2}{2} & \text{if } |a-b|<1\\ (|a-b|-\frac{1}{2})& \text{if } |a-b|\geq 1 \end{cases}\]  the Huber loss function.
			The prediction loss \(L_{\text{pred}}\) addresses the dynamics in the original space:
			\begin{equation*}
				L_{\text{pred}}
				=\mathcal{L}_{\text{Huber}}\left(\Psi(z'),x'\right).
			\end{equation*} 
			The dynamics loss \(L_{\text{dyn}}\) enforces linear dynamics in the feature space: 
			\begin{equation}\label{loss:dyn}
				L_{\text{dyn}}=\mathcal{L}_{\text{Huber}}\left(z\boldsymbol{K},z'\right).
			\end{equation}
            The Bellman loss function ensures that the learned features locally satisfy the Bellman equation, thus aligning the latent representation with the underlying \ac{mdp} dynamics:
			\begin{equation}\label{loss:bell}
				L_{\text{bell}}=\mathcal{L}_{\text{Huber}}\left(z v,\boldsymbol{R}+\gamma z'v\right).
			\end{equation}
            The sparse loss \(L_{\text{spar}}\) penalizes non-active neurons via $L_1$ regularization \citep{champion2019data,brunton2016discovering}, automatically selecting the relevant features and adaptively determining the effective number of features:
			\[L_{\text{spar}}=\|z\|_1.\]
		The idea behind the sparse loss is to select the maximum \(k_{\max}\) (or minimum \(k_{\min}\)) number of features if all neurons are ``equivalently'' active (or non-active), or to select those that are more active than others (\(k_{\min}<k<k_{\max}\)).
            The orthogonal loss  \(L_{\text{orth}}\) requires that the features are linearly independent:
			\begin{equation}\label{loss:orth}
				L_{\text{orth}}=\|z^Tz-\boldsymbol{I_k}\|_F,
			\end{equation}
			where \(\|.\|_F\) represents the Frobenius norm.
		
		Finally, the \ac{kae} algorithm, used in our setting, solves the optimization problem
			\[\min_{\Phi, K,v}L_{\text{tot}}(\Phi, K,v).\]
			We note that, in addition to the features provided by the encoder \(\Phi\) and the Koopman matrix \(\boldsymbol{K}\), due to the Bellman loss \eqref{loss:bell}, the algorithm \ac{kae} described above also learns the local weight vector \(v\). However, only features are required for the subsequent steps of the algorithm \ac{kae_lspi} (see Section \ref{sec:main}). Furthermore, in our setup, the most important losses will be the  dynamics and Bellman losses \eqref{loss:dyn}-\eqref{loss:bell}, since they relate the Koopman operator framework to \ac{rl}.
		
		\section{The Koopman autoencoder-based least-squares policy iteration algorithm}
		\label{sec:main}

		Assume that we have a \ac{mdp} \((\mathcal{S},\mathcal{A},\mathcal{P},R,\gamma)\) and a fixed policy \(\pi\). 	
		The Bellman optimal equation \eqref{eq:bellman_optimal} can be written as
		\begin{equation*}
			Q^*(s,a)=\sum_{s'\in \mathcal{S}} \mathcal{P}(s'|s,a) R(s,a,s')+\gamma \max_\pi (\mathcal{K}^\pi Q^\pi)(s,a),
		\end{equation*}
		where, inspired by \citet{rozwood2024koopman}, we define the \textit{stochastic  Koopman operator} \(\mathcal{K}^\pi\) by
		
			\begin{equation}\label{eq:koopman_stochastic}
			(\mathcal{K}^\pi Q^\pi)(s,a)= \mathbb{E}_{s'\sim\mathcal{P}(.|s,a)}\left[Q^\pi(s',\pi(s'))\right]=\sum_{s'\in \mathcal{S}} \mathcal{P}(s'|s,a) Q^\pi(s',\pi(s')).
		\end{equation}

		Suppose that we have a dataset \(D=\{s_i,a_i,r_i,s'_i\}_{i=1,\ldots, L}\) and a dictionary of basis functions \(\phi=\big(\phi_1 \,\phi_2\,\ldots\,\phi_k\big)\) \(\in \mathbb{R}^{1\times k}\) where \(Q^\pi(s,a)=\phi(s,a)w^\pi\), and we want to approximate the operator \(\mathcal{K}^\pi\) by a matrix \(\boldsymbol{K^\pi}\in \mathbb{R}^{k\times k}\):
		\[(\mathcal{K}^\pi \phi)(s,a)\approx \phi(s,a) \boldsymbol{K^\pi}.\]
		From \eqref{eq:koopman_stochastic} we have, 
		\begin{eqnarray}\label{eq:koopman_stochastic_2}
			(\mathcal{K}^\pi \phi)(s,a)&=&\sum_{s'\in \mathcal{S}} \mathcal{P}(s'|s,a) \phi(s',\pi(s'))
		\end{eqnarray}
		which can be learned by using a Monte-Carlo procedure: 
		\begin{equation}\label{eq:koopman_stochastic_3}
			\dfrac{1}{L_{(s,a)}}\sum_{j=1}^{L_{(s,a)}} \phi\big(s'^{(j)},\pi\big(s'^{(j)}\big)\big),
		\end{equation}
		where \(L_{(s,a)}\) is the number of realizations for the fixed state-action pair \((s,a)\).
		The approximation $\boldsymbol{K^\pi}$ is then obtained by the following optimization problem
		{\small \begin{equation*}\label{eq:koopman_opti}
				\min_{\boldsymbol{K}}\dfrac{1}{L}\sum_{i=1}^{L} \dfrac{1}{L_{(s_i,a_i)}}\sum_{j=1}^{L_{(s_i,a_i)}}\left\|	 \phi\big(s_i'^{(j)},\pi\big(s_i'^{(j)}\big)\big)- \phi\big(s_i,a_i\big)\boldsymbol{K}\right\|^2.
		\end{equation*}}
		The solution is given by
		\[\mathbf{K^\pi=G^+A^\pi},\]
		with
			\begin{equation*}
			\mathbf{G} = \dfrac{1}{L}\sum_{i=1}^{L}\phi\big(s_i,a_i)^\top\phi\big(s_i,a_i) \text{ and }
			\mathbf{A^\pi} = \dfrac{1}{L}\sum_{i=1}^{L}\dfrac{1}{L_{(s_i,a_i)}}\sum_{j=1}^{L_{(s_i,a_i)}}\phi(s_i,a_i)^\top\phi\big(s_i'^{(j)},\pi\big(s_i'^{(j)}\big)\big).
		\end{equation*}

			Following the discussion presented in \citet[Section 6]{lagoudakis2003least}, we note that, in model-based \ac{mdp} settings, it is possible to generate multiple independent successor realizations for a fixed state-action pair \((s,a)\). In contrast, when samples are collected directly from the underlying process, repeated realizations of exactly the same \((s,a)\) are generally not available. 
			Consequently, if we only choose one realization per state-action pair \({(s,a)}\), then \eqref{eq:koopman_stochastic_3} simply becomes \(\phi(s',\pi(s'))\) and in this case, \eqref{eq:fixed_point_method} can be written as
		
			\begin{equation}\label{eq:fixed_point_method_koopman}
			\mathbf{\Phi^\top} 	\left( \boldsymbol{\Phi}-\gamma \mathbf{\Phi \boldsymbol{K^\pi} }\right) w^\pi = \mathbf{\Phi^\top R }  \Longleftrightarrow	 \mathbf{\Phi^\top \Phi}\left( \mathbf{I_k}-\gamma \boldsymbol{K^\pi} \right) w^\pi = \mathbf{\Phi^\top R}.
		\end{equation}
		This equation represents the Koopman matrix formulation of \textit{least-squares fixed-point approximation} in classical \ac{lspi} methods.
		Therefore, we can use it to derive a \ac{kae_lspi} algorithm where, in contrast to \ac{lspi} and \ac{klspi} algorithms, neither the features \(\phi\) nor a kernel \(\mathbf{k}(\cdot,\cdot)\) are \textit{a priori} fixed. 
		
		Based on the above, the \ac{kae_lstd}-Q (Algorithm \ref{alg:kae_lstd}) consists of two parts. 
		\begin{itemize}
			\item 
			First, we use the \ac{kae} to learn the dictionary of basis functions from data pairs \(\{x_i=(s_i,a_i), x_{i+1}=\big(s'_i,\pi(s_i')\big)\}_{i=1,\ldots,L}\), generated by a  policy \(\pi\). Due in particular to the dynamics loss \eqref{loss:dyn} in the \ac{kae} algorithm, the learned basis functions \(\phi_\theta\) are then constrained to satisfy the following Koopman relation 
			\begin{equation}\label{eq:koopman_constraint}
				\phi_\theta(s',\pi(s'))=\phi_\theta(s,a)\boldsymbol{K}_\theta^\pi.
			\end{equation}
			\item Secondly, we fix the learned features \(\phi\) and, according to \eqref{eq:fixed_point_method_koopman}, we use them in the classical \ac{lstd}-Q algorithm (See \citet{lagoudakis2003least}).
		\end{itemize}

\begin{algorithm2e}[ht]
	\caption{\ac{kae_lstd}-Q algorithm}
	\label{alg:kae_lstd}
	\SetKwInOut{Input}{Inputs}
	\SetKwInOut{Output}{Output}
	
	\Input{%
		Data \(D=\{s_i,a_i,r_i,s'_i,a'_i\}_{i=1,\ldots,L}\) collected with a policy \(\pi\)\\
		Discount factor \(\gamma\)\\
		Fixed minimum and maximum number of learned basis functions \(k_{\min}\) and \(k_{\max}\)
	}
	\Output{\(\hat{Q}^\pi\)}
	
	Compute \(\boldsymbol{R}\in \mathbb{R}^{L\times 1}\) from~\eqref{eq:phi_r}\;
	
	\tcp{\underline{\ac{kae} step}:}
	Apply the \ac{kae} algorithm on \(\{x_i=(s_i,a_i), x_{i}'=\big(s'_i,\pi(s'_i)\big)\}_{i=1,\ldots,L}\) to learn \(\phi_\theta=\begin{pmatrix}\phi_{\theta,1}&\phi_{\theta,2}&\ldots&\phi_{\theta,k}\end{pmatrix}\) where \(k_{\min} \leq k\leq k_{\max}\)\;
	
	Compute \(\boldsymbol{\Phi}\in \mathbb{R}^{L\times k}\) from~\eqref{eq:phi_r}\;
	
	Set \(b = \boldsymbol{\Phi}^{T}\boldsymbol{R}\in \mathbb{R}^{k\times 1}\)\;
	
	Compute \(\boldsymbol{\Phi'_{\pi}}\in \mathbb{R}^{L\times k}\) from~\eqref{eq:phi_r}\;
	
	\tcp{\underline{Policy evaluation}:}
	\(\boldsymbol{A_{\pi}} = \boldsymbol{\Phi}^{T}\bigl(\boldsymbol{\Phi} - \gamma \boldsymbol{\Phi'_{\pi}}\bigr)\in \mathbb{R}^{k\times k}\)\;
	
	\(w^{\pi_j} = \mathbf{A_{\pi}^{+}} b\in \mathbb{R}^{k\times 1}\)\;
	
	\(\hat{Q}^{\pi}(s,a) = \phi_\theta(s,a) w^{\pi}\).
	
\end{algorithm2e}

		
		It should be noted that Algorithm \ref{alg:kae_lstd} only represents a \(Q\)-learning step and therefore cannot be used for control. Adding a policy improvement step thus results in Algorithm \ref{alg:kae_lspi}.

\begin{algorithm2e}[ht]
	\caption{\ac{kae_lspi} algorithm}
	\label{alg:kae_lspi}
	\SetKwInOut{Input}{Inputs}
	\SetKwInOut{Output}{Output}
	
	\Input{%
		Data \(D_0=\{s_i,a_i,r_i,s'_i,a'_i\}_{i=1,\ldots,L}\) collected with a random policy \(\pi_0\)\\
		Discount factor \(\gamma\)\\
		Fixed minimum and maximum number of learned basis functions \(k_{\min}\) and \(k_{\max}\)\\
		Maximum number of iterations \(M\)
	}
	\Output{Approximate optimal policy \(\hat{\pi}^*\)}
	
	\For{\(j=0\) to \(M-1\)}{
		For the current dataset \(D_j\) and the policy \(\pi_j\), use Algorithm~\ref{alg:kae_lstd} to approximate \(\hat{Q}^{\pi_j}\)\;
		
		\tcp{\underline{Policy improvement}:}
		\(\pi_{j+1}(s) = \arg\max_{a\in\mathcal{A}} \hat{Q}^{\pi_j}(s,a)\)\;
		
		\If{\(\pi_{j+1} = \pi_j\)}{
			\(\hat{\pi}^* = \pi_{j+1}\)\;
			\textbf{break}\;
		}
		
		\tcp{\underline{Collect new data}}
		\(D_{j+1} = \{(s_i,a_i,r_i,s'_i,a_i')\}_{i=1,\ldots,L}\) where \(a_i'=\pi_{j+1}(s_i')\).
	}
\end{algorithm2e}

		The data regeneration step of Algorithm \ref{alg:kae_lspi} can take two forms. Either we collect all new  transitions \(\{(s_i,a_i,r_i,s'_i,a_i')\}_{i=1,\ldots,L}\) using the new policy \(\pi_{j+1}\), or  we  collect only the new next actions \(a_i'=\pi_{j+1}(s_i')\). We will use the second option for both   \ac{kae_lspi} and  \ac{klspi} since it is close to  the classical \ac{lspi}.
		
		In essence, Algorithm \ref{alg:kae_lspi} is the classical \ac{lspi} algorithm based on the \textit{least-squares fixed-point approximation} with an \emph{on-policy} learned  basis functions constrained by the relation \eqref{eq:koopman_constraint}. This approach is related to the \ac{klspi} algorithm \citep{xu2007kernel,huang2012novel}, where the \ac{ald} procedure plays the same role as \ac{kae}.

			Following the same ideas as in \citet{xu2007kernel}, we present a convergence analysis of the algorithm \ref{alg:kae_lspi} in Appendix~\ref{append:converg}. Before some illustrations, we conclude this section by noting the following two limitations of our algorithm mainly due  to the \ac{kae} step:
			\begin{itemize}
				\item \ac{kae_lspi} requires more runtime and more data,
				\item since it is an algorithm \ac{nn}, \ac{kae_lspi} is highly dependent on hyperparameters, which must be tuned.
			\end{itemize}

		\section{Numerical examples}
		\label{sec:illustrations}
		
		In this section, we first compare the \ac{lspi}, \ac{klspi}, and \ac{kae_lspi} algorithms on the discrete \ac{mdp} chain walk problem (with \(20\) and \(50\) states). In this example, the comparison will focus on the number of features in the dictionary, convergence to an optimal or near-optimal policy, the difference between the final and the optimal policy, and the convergence iteration step. Secondly, we compare the \ac{lspi}, \ac{klspi} and \ac{kae_lspi} results for the continuous inverted pendulum control problem
		 using the performance of the control policies as a function of the number of training episodes \citep{lagoudakis2003least}.
		\subsection{Chain walk problem}
		
		This \ac{mdp} problem, studied in \citet{lagoudakis2003least,xu2007kernel}, consists of a chain with \(n\) states (\(n=20\) or \(50\) in our cases): \(\mathcal{S}= \{1,2,\ldots,n\}\). In each state, there are two possible actions, ``left'' (\(L\)) and ``right'' (\(R\)) indicating where to move: \(\mathcal{A}= \{L,R\}\). In both algorithms, the two actions are encoded as \(1\) and \(2\), respectively. The transition model is given by
		\[ \begin{cases}
			\mathcal{P}\big(\max\{1,s-1\}|s,L\big)=0.9=\mathcal{P}\big(\min\{n,s+1\}|s,R\big)\\
			\mathcal{P}\big(\min\{n,s+1\}|s,L\big)=0.1=\mathcal{P}\big(\max\{1,s-1\}|s,R\big),
		\end{cases}\]
		i.e., each action proceeds in the intended direction with probability \(0.9\) and fails with probability \(0.1\). The discount factor is given by \(\gamma=0.9\) and we will give the reward function for each case afterward.

		For \ac{lspi} algorithm, the fixed dictionary  is set	according to \citet{lagoudakis2003least}. Also, for \ac{klspi} algorithm features are learned from the \ac{ald} procedure by using a \ac{rbf} kernel as in \citet{xu2007kernel}. For more details, we encourage the reader to see   Appendix \ref{append:lspi_klspi}. 
        Finally, for Algorithm \ref{alg:kae_lspi}, we used the Z-score normalization on the data \(x\) and \(x'\) and the hyperparameters summarized in Table \ref{tab:kae} in Appendix \ref{append:parameters}.
		
		\subsubsection{First case (\(n=20\))} The reward function is given by: 
			\(r(1)=r(20)=1\text{ and }
			r(s)=0, \text{ otherwise}.\)
			Following \citet{lagoudakis2003least}, for both algorithms, the initial dataset consists of \(5000\) transitions collected along a single trajectory generated  by running the model with random actions.	
			The optimal policy is to move left in states \(1-10\) and right in states \(11-20\). 
			The results of the three algorithms are reported in Table \ref{tab:20states} as the mean $\pm$ standard deviation over \(50\) independent runs, with a \(95\%\) \ac{ci}. 
				Although  the number of features is fixed for \ac{lspi}, 
				for \ac{klspi} and \ac{kae_lspi}, this quantity is determined during the
				final iteration of each run and may vary across runs.
				The convergence rate reports the fraction of the \(50\) runs that reached a fixed point
				($\pi_{j+1} = \pi_j$) before the maximum number of iterations (set here to \(7\)). It measures the stability of the algorithms.
				The optimal policy match rate represents the fraction of \(50\) runs whose final policy is exactly $\pi^*$.
				The final policy mismatch rate is the percentage of states for which the final policy
				differs from $\pi^*$. Its aims to measure the quality of the near-optimal solution.
				Finally, the convergence iteration corresponds to the number of \ac{lspi} iterations performed before convergence, if convergence occurs.

			\begin{table}[h]
				\centering
				\caption{Comparison of \ac{lspi}, \ac{klspi}, and \ac{kae_lspi} on the \(20\)-state chain walk problem.}
				\label{tab:20states}
				\begin{tabular}{||p{5.3cm}|p{2cm}|p{2.2cm}|p{2cm}||}
					\hline
					& \ac{lspi} & \ac{klspi} & \ac{kae_lspi} \\
					\hline
					Number of features              & $8$              & $40.00 \pm 0.00$      &  $16.00 \pm 0.88$  \\
					\hline
					Convergence rate                & $94\%$           & $100\%$         & $100\%$             \\
					\hline
					Optimal policy match rate       & $94\%$           & $98\%$         &  $96\%$                \\
					\hline
					Final policy mismatch rate (\%) & $1.50 \pm 6.00$ & $0.10 \pm 0.71$ & $0.20 \pm 0.99$                 \\
					\hline
					Convergence iteration           & $3.36 \pm 1.26$  & $2.06 \pm 0.24$ &    $3.20\pm 0.83$               \\
					\hline
				\end{tabular}
			\end{table}

			Figs.\ \ref{fig:20_state_chain_lspi}-\ref{fig:20_state_chain_kae_lspi} in Appendix \ref{append:policy} show the policy iteration results obtained with the algorithms for one typical run.

			\subsubsection{Second case (\(n=50\))} The reward function is given by: 
				\(r(10)=r(41)=1\text{ and }
				r(s)=0, \text{ otherwise}.\)
				The initial dataset consists of \(10000\) transitions collected along a single trajectory generated  by running the model with random actions. 
					The optimal policy is to move right in states \(1-9\) and \(26-41\) and left in states \(10-25\) and \(42-50\). As previously, the results of the three algorithms are reported in Table \ref{tab:50states} and
						Figs.\ \ref{fig:50_state_chain_lspi}-\ref{fig:50_state_chain_kae_lspi} in Appendix \ref{append:policy} illustrate the policy iteration results obtained for a typical run.
					
					\begin{table}[h]
						\centering
						\caption{Comparison of \ac{lspi}, \ac{klspi}, and \ac{kae_lspi} on the \(50\)-state chain walk problem.}
						\label{tab:50states}
						\begin{tabular}{||p{5.3cm}|p{2cm}|p{2.2cm}|p{2cm}||}
							\hline
							& \ac{lspi} & \ac{klspi} & \ac{kae_lspi} \\
							\hline
							Number of features              & $22$              & $100.00 \pm 0.00$      & $26.30\pm1.90$ \\
							\hline
							Convergence rate                & $90\%$           & $100\%$         & $88\%$             \\
							\hline
							Optimal policy match rate       & $0\%$           & $30\%$         &    $38\%$              \\
							\hline
							Final policy mismatch rate (\%) & $7.80 \pm 12.98$ & $1.92 \pm 1.51$ &   $1.88 \pm 1.78$               \\
							\hline
							Convergence iteration           & $5.33 \pm 1.58$  & $2.66 \pm 0.48$ &         $4.75 \pm 2.16$         \\
							\hline
						\end{tabular}
					\end{table}

					Regarding this chainwalk problem, we can conclude that \ac{kae_lspi} performs just as well as the other two algorithms when learning features from the data. Finally, note that in this example, the state and action spaces were finite discrete spaces. The next example will be more challenging in terms of the state space.

					\subsection{Control inverted pendulum on a cart}

					\begin{figure}[h!]
						\centering
						
						\begin{minipage}[b]{0.33\textwidth}
							\centering
							\includegraphics[width=\linewidth]{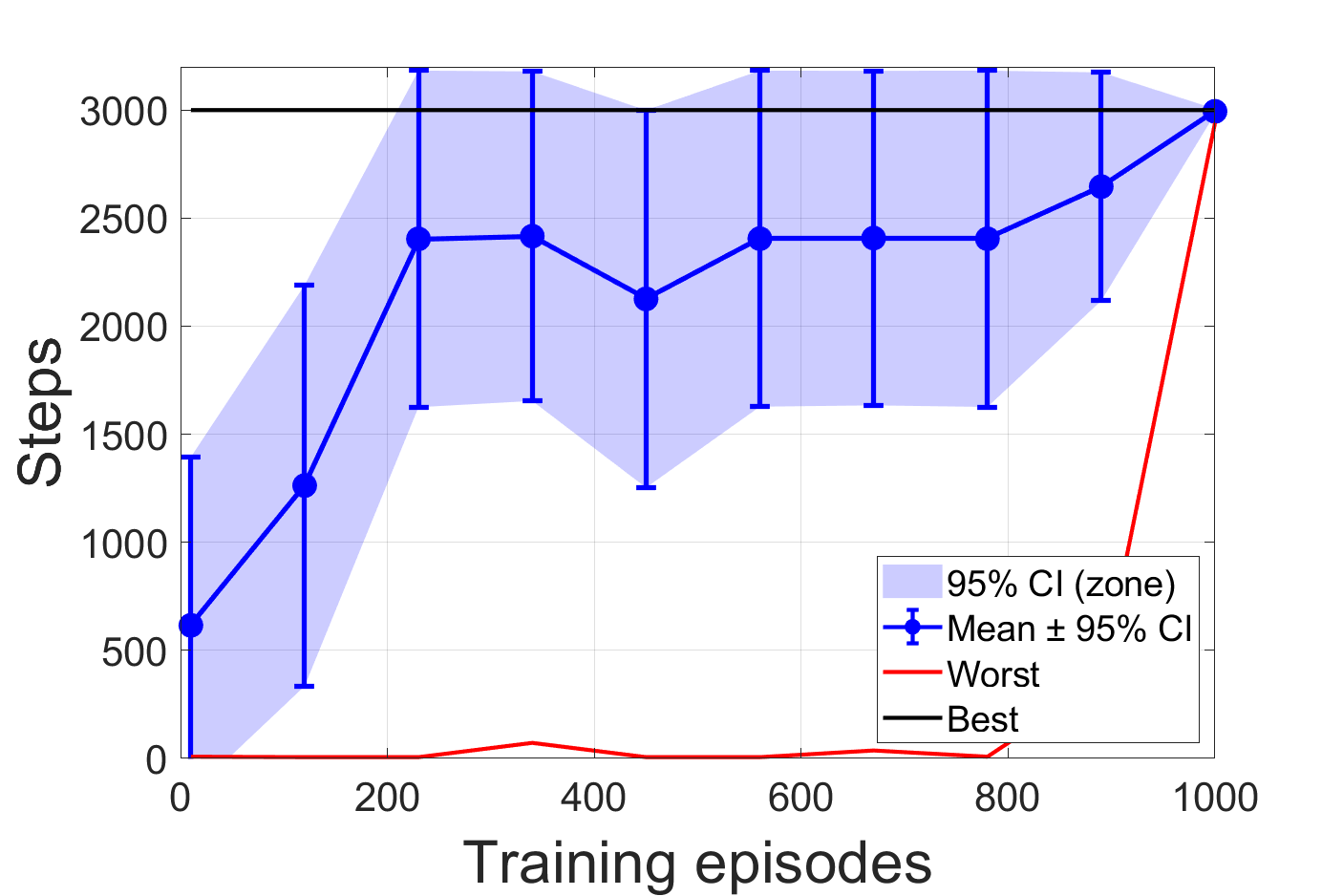}
							\refstepcounter{subfigure}\label{fig:pendulum_lspi}%
							\small (a)
						\end{minipage}\hfill
						\begin{minipage}[b]{0.33\textwidth}
							\centering
							\includegraphics[width=\linewidth]{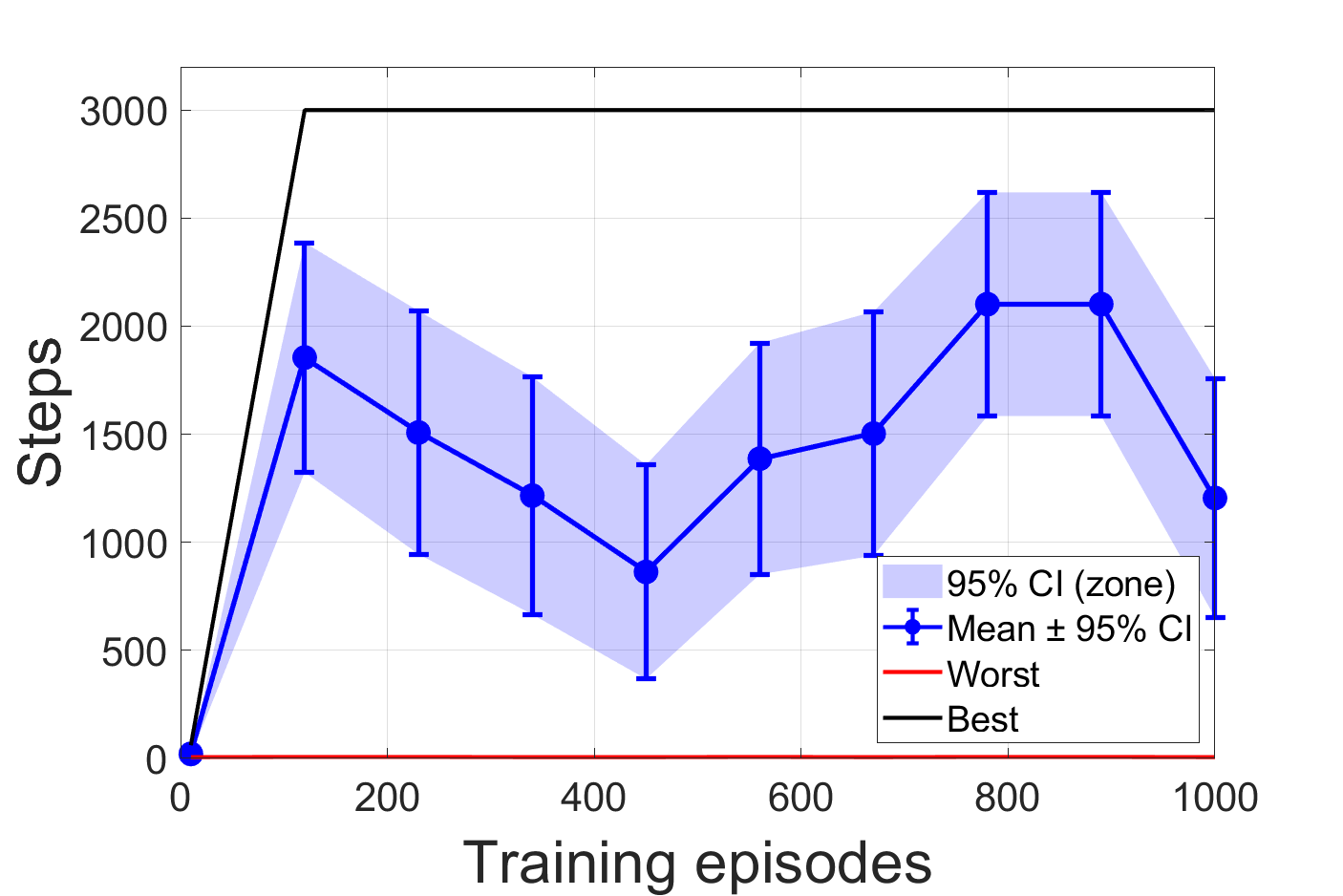}
							\refstepcounter{subfigure}\label{fig:pendulum_klspi}%
							\small (b)
						\end{minipage}\hfill
						\begin{minipage}[b]{0.33\textwidth}
							\centering
							\includegraphics[width=\linewidth]{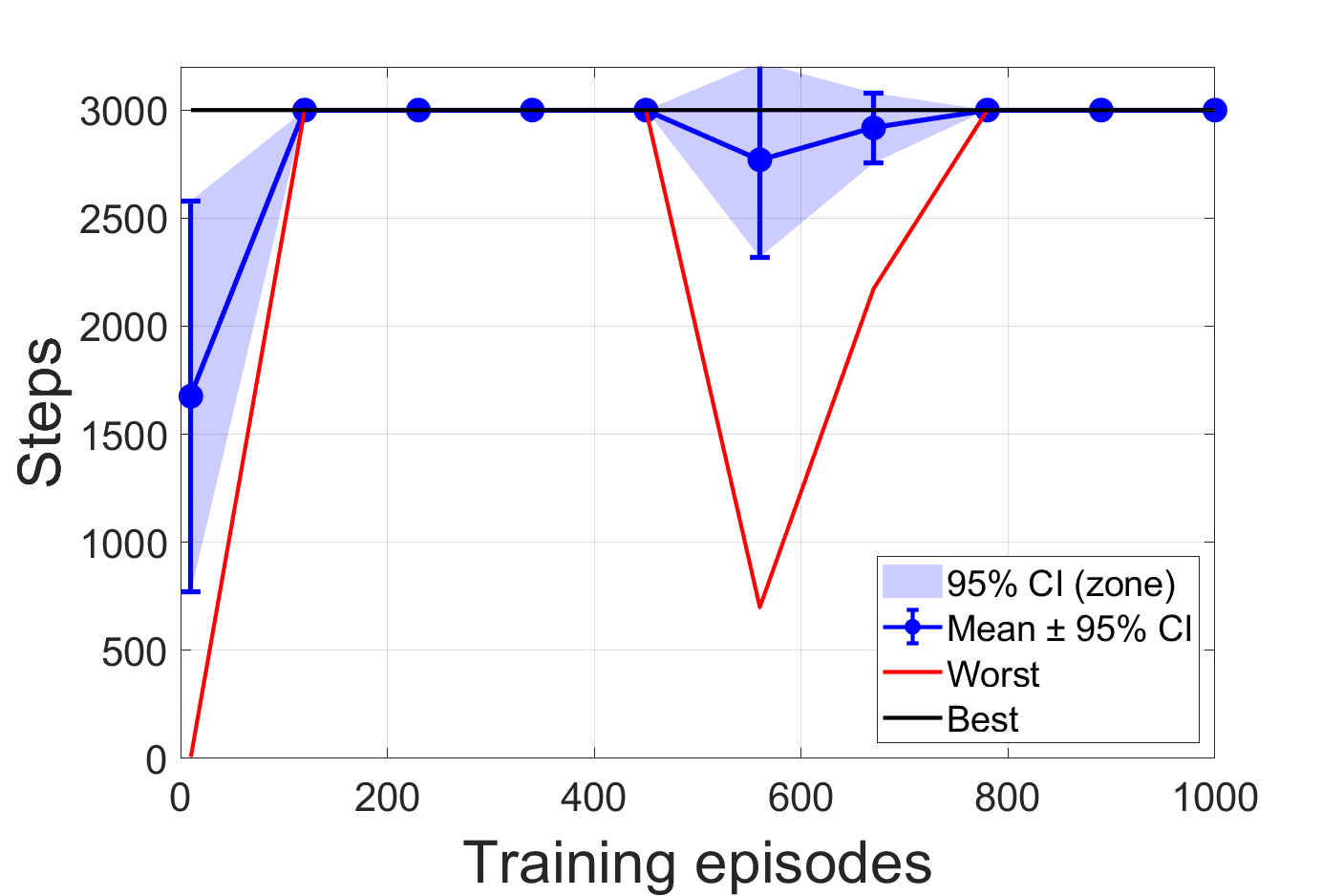}
							\refstepcounter{subfigure}\label{fig:pendulum_kae_performance}%
							\small (c)
						\end{minipage}
						
						\caption{(a) \ac{lspi} performance. (b) \ac{klspi} performance. (c) \ac{kae_lspi} performance.}
						\label{fig:pendulum}
					\end{figure}
                    
					For this problem we use the same setup as in \citet{lagoudakis2003least}. The inverted pendulum on a cart (or cart-pole system) is described by the nonlinear
					dynamics 	depending on the current state \(s=(\theta,\dot \theta)\in \mathcal{S}=\mathbb{R}^2\) and the current
					(noisy) control \(a\):
					\begin{equation}\label{eq:pendulum}
						\ddot{\theta}=\dfrac{g\sin(\theta)-\alpha m\ell (\dot \theta)^2\sin(2\theta)/2-\alpha \cos(\theta)a}{4\ell/3-\alpha m\ell \cos^2(\theta)},
					\end{equation}
					where \(\theta\) and \(\dot \theta\) are vertical angle and angular velocity, \(g = 9.8m/s^2\) is the gravity constant, \(m= 2.0kg\) is the mass of the pendulum, \(M= 8.0 kg\) is the mass of the cart, \(l = 0.5 m\) is the length of the pendulum,
					and \(\alpha =1/(m +M)\). The goal is to maintain the pendulum in the horizontal position (\(\vert\theta\vert < \pi/2\)) for a maximum of \(3000\) steps. The agent uses three actions: 
					\(\mathcal{A}= \{-50,0,+50\}\) with a uniform noise in \([-10, 10]\) added to the chosen
					action. The discount factor for this \ac{mdp} problem
					is set to \(\gamma=0.95\) and the reward function is given by
                    \[r(s)= 0 \text{ if } \vert\theta\vert\leq \pi/2 \text{ and } r(s)=-1 \text{ otherwise.}\]
					The time step for Equation \eqref{eq:pendulum} was set to \(0.1\) seconds. 
					During the training, for both algorithms, the data samples were collected by solving the dynamics \eqref{eq:pendulum} via RK4 method with a random policy and ten different numbers of episodes ranging from \(10\) to \(1000\). Each episode starts from an initial condition very close to the equilibrium point \((0,0)\) and ends when \(\vert\theta\vert> \pi/2\) or the simulation reaches \(20\) steps. During testing, \(200\) initial conditions near \((0,0)\) were simulated, with each simulation ending when \(\vert\theta\vert> \pi/2\) or after \(3000\) steps.
					Moreover, the algorithm's performance was averaged over \(10\) learning runs for each episode count.
					No noise was applied to the chosen actions in the testing phase.
					
					Following \citet{lagoudakis2003least}, for the \ac{lspi} algorithm, as in \eqref{eq:rbf} (see Appendix \ref{append:lspi_klspi}), we set \(30\) \ac{rbf}s 
					{\small \begin{equation*}\label{eq:rbf_2}
							\phi(s,a)=\left(
							I(a=a_i)\times 1, \ldots , I(a=a_i)\times \exp\left(-\frac{\|s-\mu_{9}\|^2}{2\sigma^2}\right) \right),\end{equation*}}
					for \(a_i\in \{-50,0,+50\}\), where \(\sigma=1\) and the \(\mu_j\)'s are the \(9\) points of the grid \(\{-\pi/4,0,\pi/4\}\times \{-1,0,+1\}\).
					
						For the \ac{kae_lspi} algorithm, we set the hyperparameters \(k_{\min}=22, k_{\max}=35\), and the hidden layers of the encoder and decoder to \([128,64]\) and	 \([64,128]\), respectively, the learning rate to \(10^{-4}\), and the Bellman loss weight (\(\lambda_{\text{bell}}\)) to \(5\). All other parameters remain the same as in the \(50\) states chainwalk example (see Table \ref{tab:kae} in Appendix \ref{append:parameters}).
					The performance results on the average balancing steps of the \(10\) runs in each episode are represented in Figs.\ \ref{fig:pendulum_lspi}-\ref{fig:pendulum_kae_performance}, with a \(95\%\) \ac{ci}.					
					Figs.\ \ref{fig:pendulum_lspi}-\ref{fig:pendulum_kae_performance} show that, the \ac{kae_lspi} algorithm achieves better results than the other algorithms for this problem.

					\section{Conclusion and perspectives}
					\label{sec:conclusion}
					
					In this paper, we introduced the \ac{kae_lspi} algorithm which aims to learn the dictionary of features from data samples and solve \ac{rl} problems in the value-function approximation context. We discussed the convergence analysis of  \ac{kae_lspi} and demonstrate its performances through two specific examples, the chain walk problem and the controlled inverted pendulum.
					
					Many future research directions starting from the \ac{kae_lspi} algorithm are possible. We could automate the selection of hyperparameters in \ac{kae}, which is currently done manually. It would also be appropriate to apply a similar Koopman approach in the context of \ac{lsvi}. A key future investigation from the \ac{kae_lspi} idea would be to derive regret bounds in \ac{rl} from established error bounds in the Koopman framework.

					
					\acks{The authors acknowledge funding from the European Research Council (ERC Starting Grant ``KoOpeRaDE'') under the European Union's Horizon 2020 research and innovation program (Grant agreement No. 101161457). \\
						The authors also thank Michael Skowronek for discussions around the \ac{kae} implementation. }
					
					
					\section*{DECLARATION OF GENERATIVE AI AND AI-ASSISTED TECHNOLOGIES IN THE WRITING PROCESS}
					
					During the preparation of this work the authors used Deepseek AI 
					in order to translate Matlab codes 
					into Python, where the Adam optimizer was implemented for the optimization problem in the \ac{kae} technique. After using this tool/service, the authors reviewed and edited the content as needed and take full responsibility for the content of the publication.
					%


					\bibliography{christian_references_acc}
					%
					%
					%
					%
					%
					
					


					\appendix
					
					\section{Convergence analysis of Algorithm \ref{alg:kae_lspi}}\label{append:converg}
					
					We first note that the lemmas and proposition used here are closely related to those employed in \citet{xu2007kernel}. The main difference is that, unlike the \ac{klstd}-\(Q\) algorithm, in which the features are selected from the data set using a fixed kernel and the \ac{ald} procedure, the \ac{kae_lstd}-\(Q\) algorithm learns the features through the \ac{kae} procedure.

					The following lemma is a classical result in the \ac{lstd} \(Q\) learning context.	 In particular, we can apply it to the \ac{kae_lstd}-\(Q\)  algorithm.
						\begin{lemma}[Inspired by \citet{xu2007kernel,tsitsiklis1996analysis}]\label{lemma:xu}
							Let \(w^{\pi}\) denote the weight vector obtained in \ac{kae_lstd}-\(Q\), and let \(Q^{\pi}\) denote the true state-action value function of the Markov chain. Let \(\mathcal{X}=\mathcal{S}\times\mathcal{A}\) denote the state-action space. Assume that the following conditions hold:
							\begin{enumerate}
								\item[A1)] The Markov chain \(\{x_i\}\), whose states are formed by the state-action pairs of an \ac{mdp} under a stationary policy, is ergodic, with transition probability matrix \(P\). There exists a unique distribution \(\mu\) satisfying
								\(    \mu P^{T}=\mu,
								\)
								with \(\mu(x)>0\) for all \(x\in\mathcal{X}\). Depending on the cardinality of \(\mathcal{X}\), \(\mu\) may be a finite- or infinite-dimensional vector.
								\item[A2)] The transition rewards \(r(x,x')\) satisfy
								\begin{equation*}
									\mathbb{E}_{0}\left[r^{2}(x,x')\right]<\infty,
								\end{equation*}
								where \(\mathbb{E}_{0}[\cdot]\) denotes the expectation with respect to the distribution \(\mu\).
								\item[A3)] For every \(i=1,2,\ldots,k\),
								\begin{equation*}
									\mathbb{E}_{0}\left[\phi_{\theta,i}^{2}(x)\right]<\infty.
								\end{equation*}
							\end{enumerate}
							Then, the following relation holds:
							\begin{equation}\label{eq:Q_bound}
								\|\phi_\theta w^{\pi}-Q^{\pi}\|_{\Delta_\mu}
								\leq
								\frac{1}{1-\gamma}
								\|\boldsymbol{\Pi}_\theta Q^{\pi}-Q^{\pi}\|_{\Delta_\mu},
							\end{equation}
							where
							$
							\Delta_\mu=\operatorname{diag}\{\mu_i\},
							\,
							\|X\|_{\Delta_\mu}=\sqrt{X^{T}\Delta_\mu X},
							$
							and
							$
							\boldsymbol{\Pi}_\theta
							=
							\phi_\theta
							\left(\phi_\theta^{T}\Delta_\mu\phi_\theta\right)^{-1}
							\phi_\theta^{T}\Delta_\mu
							$
							denotes the \(\Delta_\mu\)-weighted projection onto the span of \(\phi_\theta\).
					\end{lemma}
					
					Similar to \citet{xu2007kernel}, it follows from Lemma \ref{lemma:xu} that the \ac{kae_lstd}-\(Q\) approximation satisfies the bound in \eqref{eq:Q_bound}. The right-hand side of \eqref{eq:Q_bound} depends on the ability of the feature space to approximate the exact \(Q^\pi\) function. In our setting, under appropriate assumptions on \(Q^\pi\), the \ac{uat} from \ac{nn} theory guarantees the existence of a \ac{nn} (and hence a corresponding set of features) capable of approximating \(Q^\pi\) arbitrarily well. The same reasoning applies to the kernel-based context, where the universality property also holds.   Furthermore, we note that, due to the orthonormality loss \eqref{loss:orth} used in the \ac{kae} procedure, the features \(\phi_{\theta,i}\) are linearly independent.
					
				The next lemma is also a classical result concerning the convergence of approximate policy-iteration algorithms.
					\begin{lemma}[\citet{lagoudakis2003least}, inspired by \citet{bertsekas1996neuro}]\label{lemma:lagoud}
							Let \(\pi_0\),
							\(\pi_1\), \(\pi_2\),\(\ldots\), \(\pi_m\) be the sequence of policies generated by the \ac{kae_lspi} iterations, and let \(\hat{Q}^{\pi_0}\),\(\hat{Q}^{\pi_1}\),\(\hat{Q}^{\pi_2}\), \(\ldots\), \(\hat{Q}^{\pi_m}\) denote the corresponding approximate value functions. Let \(\epsilon\) be a positive scalar that uniformly bounds the errors between the approximate and true value functions over all iterations:
							\begin{equation*}
								\forall m=1,2,\ldots,\quad
								\|\hat{Q}^{\pi_m}-Q^{\pi_m}\|_\infty\leq\epsilon.
							\end{equation*}
							Then, this sequence eventually produces policies whose performance is at most a constant multiple of \(\epsilon\) away from the optimal performance:
							\begin{equation*}
								\limsup_{m\to\infty}
								\|\hat{Q}^{\pi_m}-Q^{*}\|_\infty
								\leq
								\frac{2\gamma\epsilon}{(1-\gamma)^2}.
							\end{equation*}
					\end{lemma}
					
                    Similar to the discussion in \citet{lagoudakis2003least}, Lemma \ref{lemma:lagoud} implies that \ac{kae_lspi} is stable and that reducing the bound \(\epsilon\) improves the quality of the resulting policy. Two factors influence the bound \(\epsilon\): the distribution of the samples and the choice of features. In our setting, the features are learned directly from the data and, according to Lemma \ref{lemma:xu}, their approximation quality depends on the representational capacity of the \ac{nn} features space.
					
					We finally state the following convergence result for \ac{kae_lspi}, whose statement and proof are similar to Theorem 3.1 in \citet{xu2007kernel}.
						\begin{proposition}
							If the initial data samples
							\[
							\left\{\left(s_i,a_i,r_i,s_i',a_i'\right)\right\}
							\]
							are generated by an \ac{mdp} under a stationary initial policy, then the policies generated by the \ac{kae_lspi} algorithm converge to a region of the policy space characterized by suboptimal performance bounds determined by the approximation error of \ac{kae_lstd}-\(Q\) in \eqref{eq:Q_bound}. Furthermore, if the approximation error vanishes, \ac{kae_lspi} converges to an optimal policy of the \ac{mdp}.
					\end{proposition}

                   \section{Chain walk experiment}
                   \subsection{\ac{lspi} and \ac{klspi} features }\label{append:lspi_klspi}
                   For \ac{lspi} algorithm, the fixed dictionary  is set	according to \citet{lagoudakis2003least}:
                   \begin{itemize}
                       \item for the \(n=20\) case, we choose \(8\) polynomial basis functions: 
		\begin{equation*}
				\phi(s,a)=\big(
				I(a=L)\times 1,\ldots,	I(a=L)\times s^4,\,	 I(a=R)\times 1,\ldots,\, I(a=R)\times s^4\big),
		\end{equation*}
			where \(I\) is the indicator function\footnote{For example, \(\phi(s,L)=(	1,\, s,\, s^2,\, s^3,\, s^4,\,0,\, 	0,\, 0,\, 0,\, 0)\). }.
		\item for the \(n=50\) case, we choose \(22\) \ac{rbf} features:
				{\small \begin{equation}\label{eq:rbf}
							\phi(s,a)=\left(
							I(a=a_i)\times 1, \ldots , I(a=a_i)\times \exp\left(-\frac{\|s-\mu_{10}\|^2}{2\sigma^2}\right)\right)
					\end{equation}}
					where \(a_i\in \{L,R\}\), \(\sigma=4\) and \(\mu_j=1+49(j-1)/9\) for \(j=1,\ldots,10\). 
					
                   \end{itemize}

                   Also, for the \ac{klspi} algorithm, according to \citet{xu2007kernel}, we set the \ac{ald} threshold \(\mu=0.001\) and the kernel is an \ac{rbf} kernel with width \(\sigma=0.4\):
		\[\mathbf{k}(z_1,z_2)=\exp\left(-\frac{\|z_1-z_2\|^2}{2\sigma^2}\right).\]
		Moreover, the data samples were rescaled such that \(s\in \{1/n,2/n,\ldots,n/n\}\) and \(a\in\{0.1,0.2\}\).

        \subsection{\ac{kae} hyperparameters}\label{append:parameters}

                   \begin{table}[h]
			\centering
			\caption{\ac{kae} hyperparameters for the chain walk problem}
			\label{tab:kae}
			\begin{tabular}{||p{6cm}|c|c||}
				\hline
				& 20 States & 50 States \\
				\hline
				Learned features $(k_{\min},k_{\max})$
				&  $(10,17)$ & $(22,35)$ \\
				\hline
				Encoder layers  
				& $[128,64]$ & $[256,128]$ \\
				\hline
				Decoder layers  
				& $[64,128]$ & $[128,256]$ \\
				\hline
				Epochs	& $400$ & $500$ \\
				\hline
				$ \big(\lambda_{\text{red}},\, \lambda_{\text{pred}},\, \lambda_{\text{dyn}},\lambda_{\text{bell}},\lambda_{\text{spar}},\lambda_{\text{orth}}\big)$	&  \multicolumn{2}{c||}{$ \big(1,\, 1,\, 2,2,\, 0.05,\, 0.1\big)$} \\
				\hline
				Sparsity threshold 	&  \multicolumn{2}{c||}{$ 0.005$} \\
				\hline
				Learning rate 	& \multicolumn{2}{c||}{$10^{-6}$} \\
				\hline
				Batch size	& \(256\) & \(512\) \\
				\hline
			\end{tabular} 
		\end{table}
		\newpage
                   \subsection{Policy iteration illustration}\label{append:policy} 
                   	\begin{figure}[h!]
			\centering
			\begin{minipage}[b]{0.3\textwidth}
				\centering
				\includegraphics[width=\linewidth]{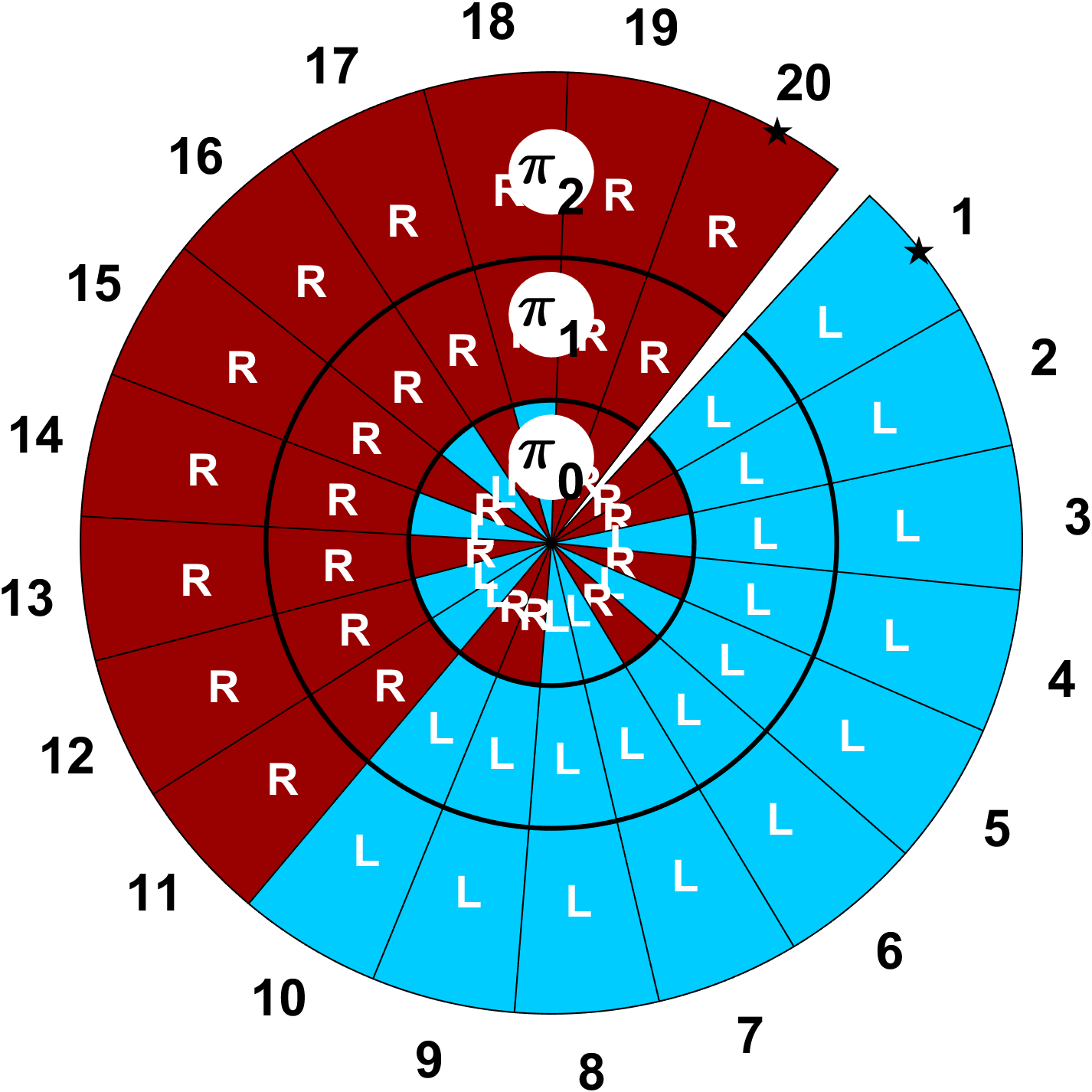}
				\refstepcounter{subfigure}\label{fig:20_state_chain_lspi}%
				\small (a)
			\end{minipage}\hfill
			\begin{minipage}[b]{0.3\textwidth}
				\centering
				\includegraphics[width=\linewidth]{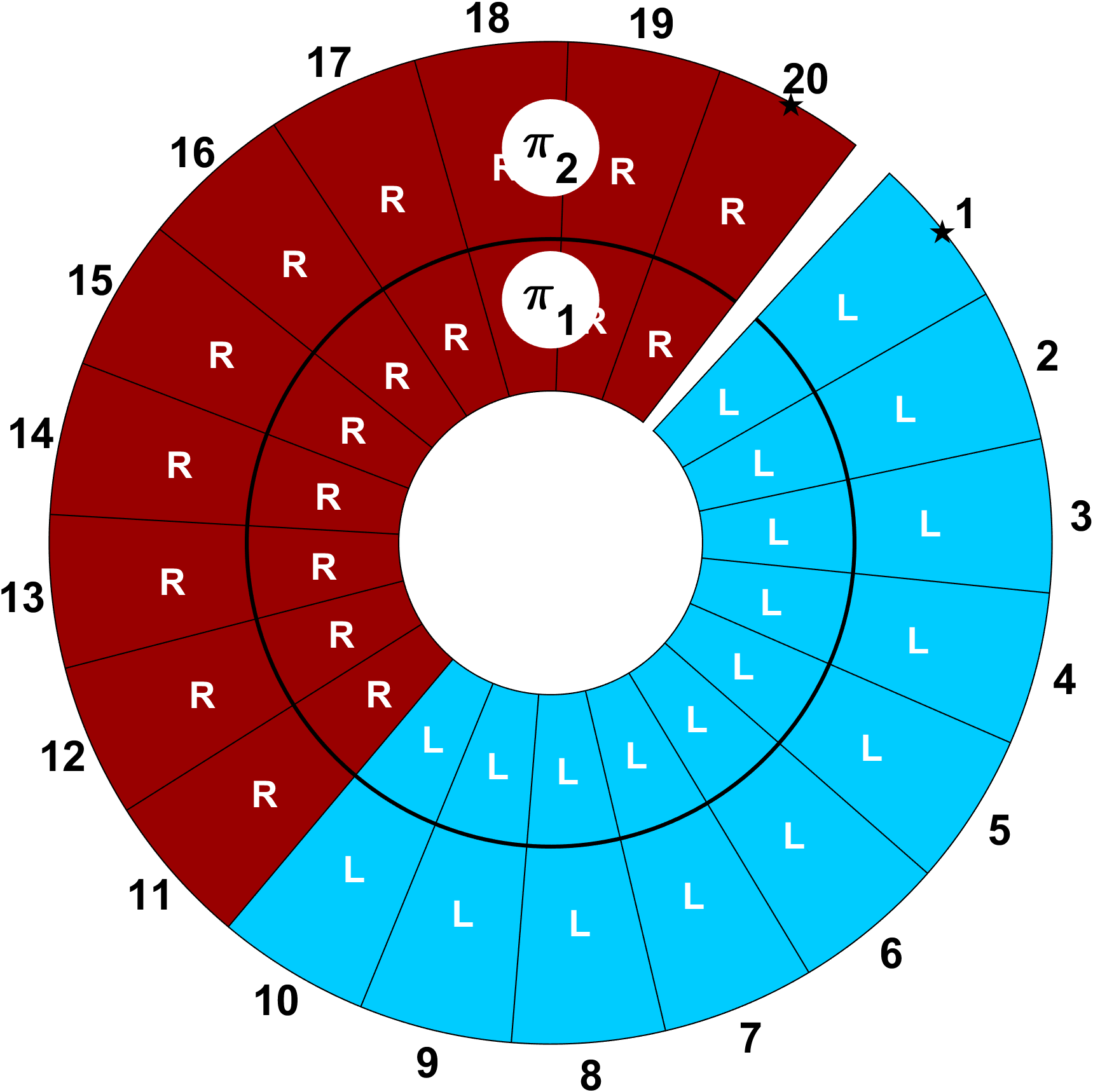}
				\refstepcounter{subfigure}\label{fig:20_state_chain_klspi}%
				\small (b)
			\end{minipage}\hfill
			\begin{minipage}[b]{0.3\textwidth}
				\centering
				\includegraphics[width=\linewidth]{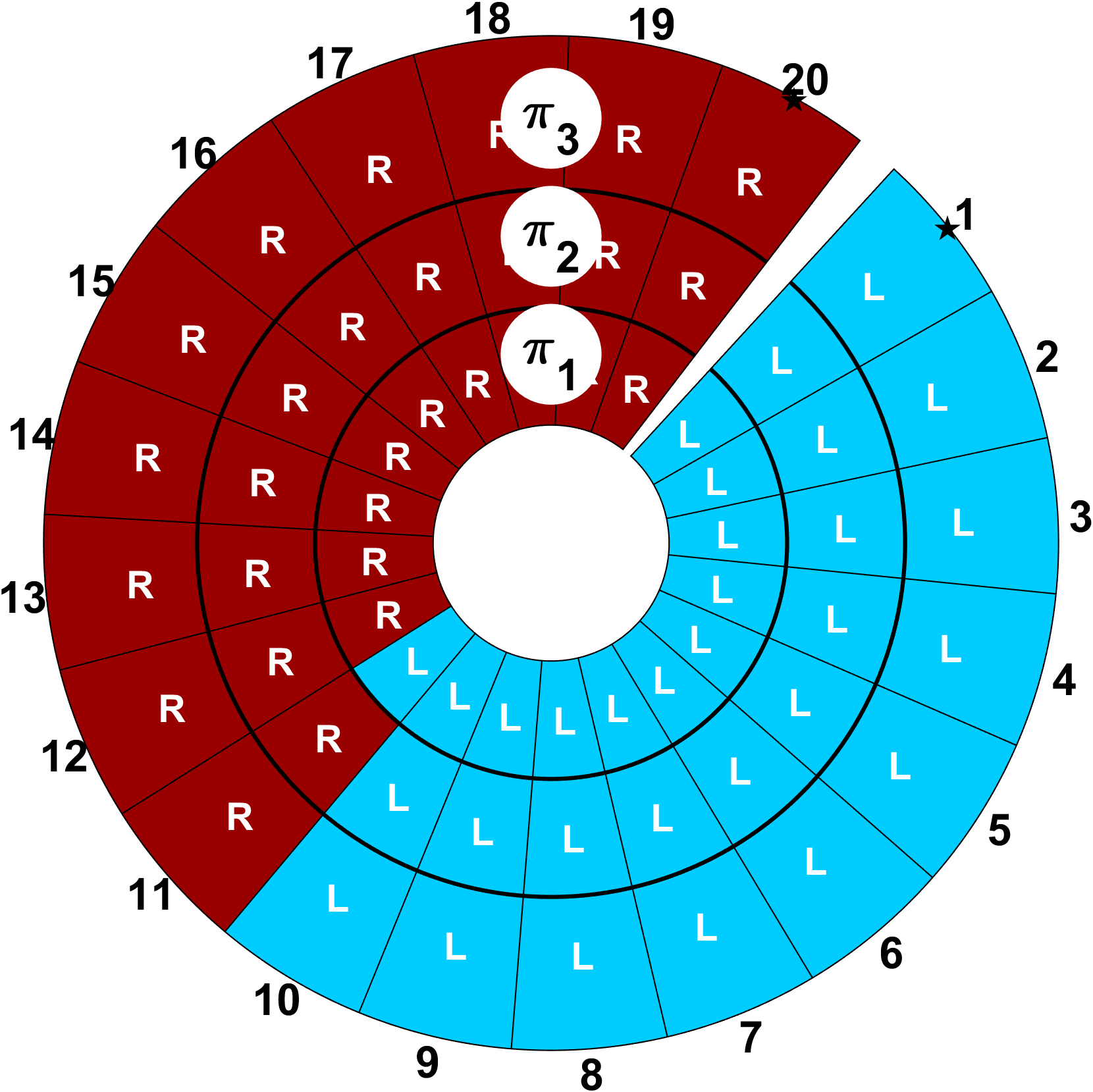}
				\refstepcounter{subfigure}\label{fig:20_state_chain_kae_lspi}%
				\small (c)
			\end{minipage}
			
			\caption{Chain walk ($n=20$) policy evolution across iterations for (a) \ac{lspi}.\, (b) \ac{klspi}.\, (c) \ac{kae_lspi}. Policies are represented in the radial direction. The white gap between states \(1\) and \(20\) indicates the absence of transitions between them.}
			\label{fig:20_state_chain}
		\end{figure}

                    	\begin{figure}[h!]
						\centering
						\begin{minipage}[b]{0.3\textwidth}
							\centering
						\includegraphics[width=\linewidth]{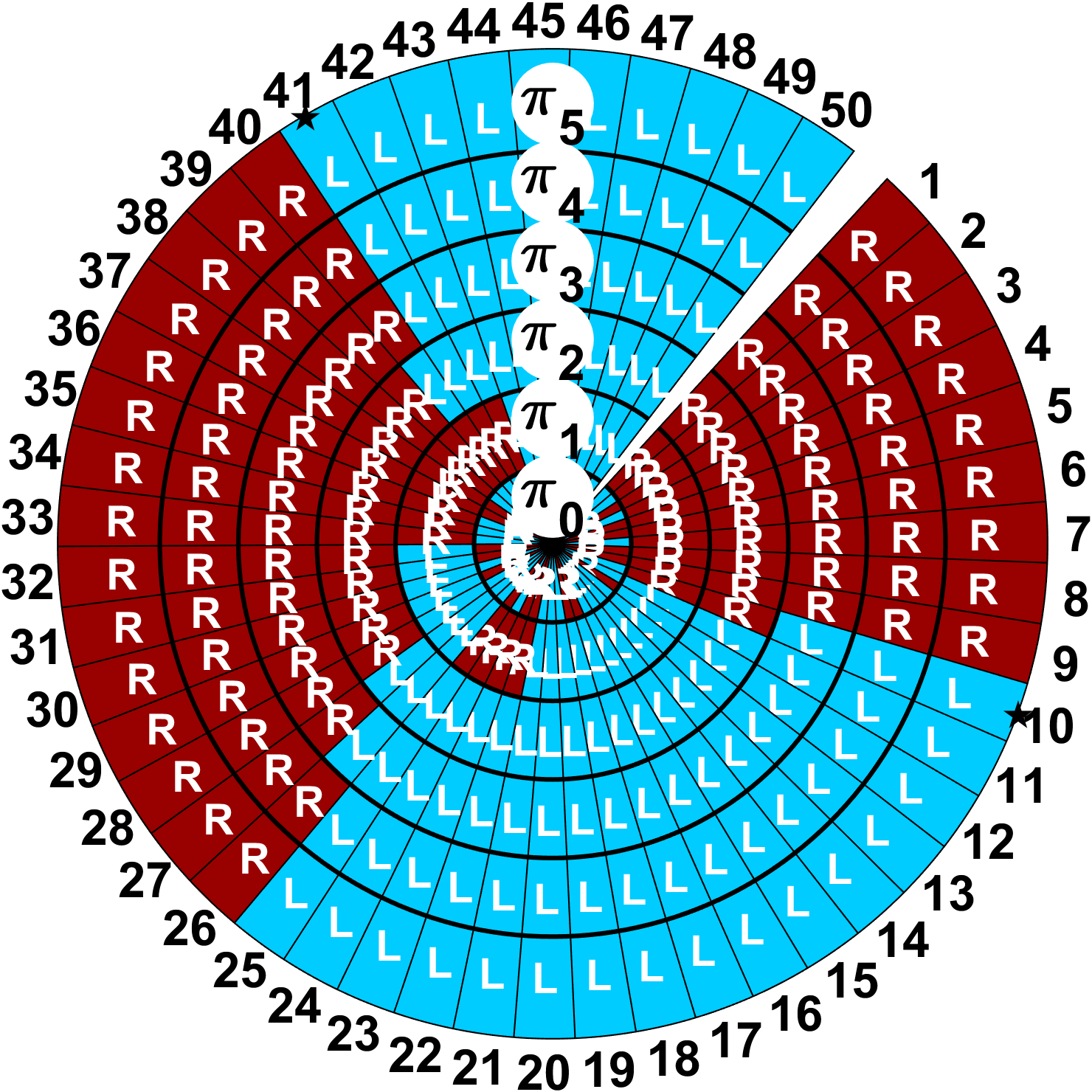}
	\refstepcounter{subfigure}\label{fig:50_state_chain_lspi}%
							\small (a)
						\end{minipage}\hfill
						\begin{minipage}[b]{0.3\textwidth}
							\centering
							\includegraphics[width=\linewidth]{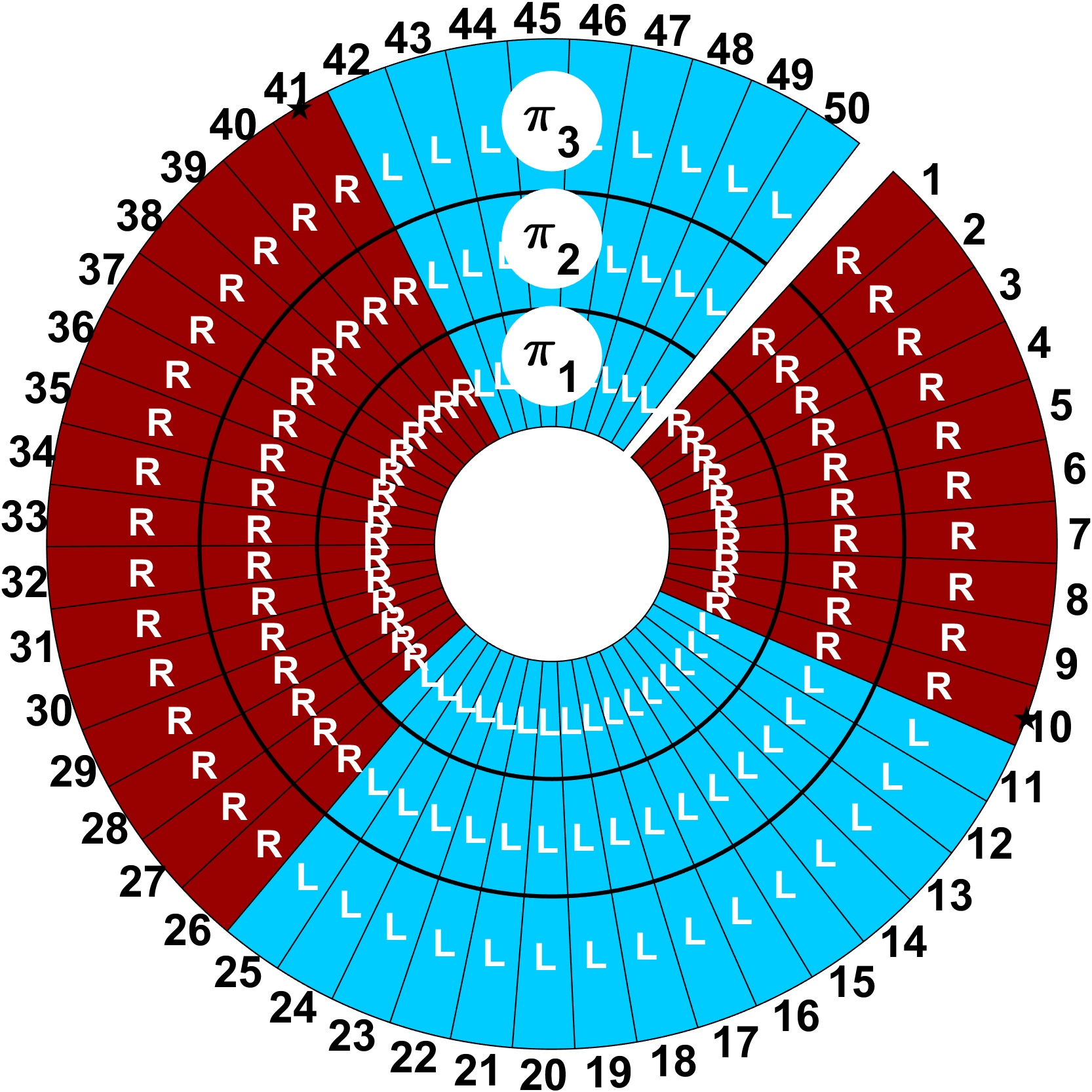}
	\refstepcounter{subfigure}\label{fig:50_state_chain_klspi}%
							\small (b)
						\end{minipage}\hfill
						\begin{minipage}[b]{0.3\textwidth}
							\centering
							\includegraphics[width=\linewidth]{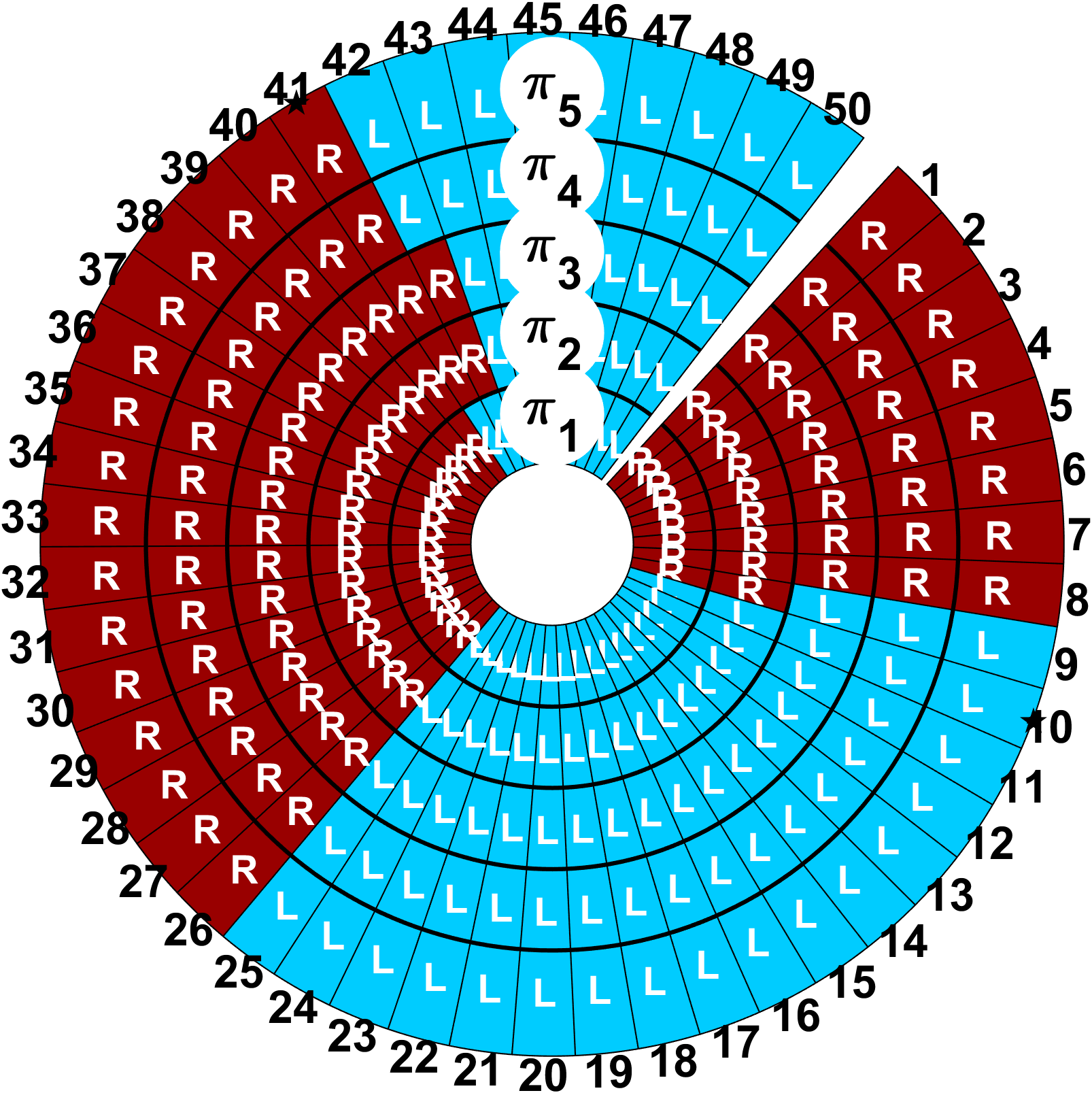}
\refstepcounter{subfigure}\label{fig:50_state_chain_kae_lspi}%
							\small (c)
						\end{minipage}
						
						\caption{Chain walk ($n=50$) policies by iteration for (a) \ac{lspi}.\, (b) \ac{klspi}.\, (c) \ac{kae_lspi}.}
						\label{fig:50_state_chain}
					\end{figure}
%
%
%

\end{document}